\documentclass[11pt]{article}

\usepackage[utf8]{inputenc}
\usepackage[T1]{fontenc}
\usepackage{lmodern}
\usepackage[margin=1in]{geometry}
\usepackage{graphicx}
\graphicspath{{figures/}}
\usepackage{booktabs}
\usepackage{amsmath}
\usepackage{amssymb}
\usepackage{xcolor}
\usepackage{microtype}
\usepackage{natbib}
\usepackage[colorlinks=true,allcolors=blue!55!black]{hyperref}
\hypersetup{
  pdftitle={No Detectable Change in Side-Level WER from Prompt-Level Context: A Preregistered Ablation on a Production Oral-History Corpus},
  pdfauthor={Theodore O. Cochran, Stephanie Dodson, Keith Nore},
  pdfsubject={Preregistered ablation of prompt-level context conditioning in production LLM speech transcription},
  pdfkeywords={speech recognition evaluation, prompt conditioning, contextual biasing, preregistration, null results, oral history}}
\usepackage{url}

\usepackage{etoolbox}
\AtBeginEnvironment{tabular}{\footnotesize}
\title{\bfseries No Detectable Change in Side-Level WER from Prompt-Level Context:\\ A Preregistered Ablation on a Production Oral-History Corpus}
\author{
  Theodore O. Cochran\thanks{\href{mailto:theo@ai4altruism.org}{theo@ai4altruism.org}}\\ \small AI for Altruism
  \and Stephanie Dodson\\ \small Independent Researcher
  \and Keith Nore\\ \small Independent Researcher
}
\date{}

\begin{document}
\maketitle

\begin{abstract}
Supplying context at inference time to a large multimodal model is an inexpensive lever for adapting speech transcription to a domain, and earlier results on smaller models reported large gains. This work tested that mechanism where it ships, in the prompt-conditioning layer of a production oral-history transcription tool, on a sample from its own production corpus. Full prompt-level context did not detectably change side-level word error rate (WER), and none of the four preregistered hypotheses was supported. The design was a within-item paired ablation, preregistered with the analysis code frozen by hash before the confirmatory batch was scored; two disclosed gpt-4o pilot sides had been scored earlier, during scorer development. Nineteen cassette sides, about 10.6 hours of degraded 1970s--80s interview audio, were reprocessed through the production code path under three prompt arms, crossed with two deployed commercial configurations, \texttt{gpt-4o-transcribe} and \texttt{gemini-2.5-flash}, and scored against operator-corrected verbatim references. For \texttt{gpt-4o-transcribe} the median paired difference between the full-context and no-context arms was $+0.6$ WER points, with a side-resampled interval of $[-1.1, +1.0]$; the Gemini estimates were too unstable to support a comparable negative inference. A post-hoc rerun found run-to-run pipeline variability larger than the confirmatory differences, so effects of that size cannot be resolved from one transcription per cell. An implementation audit verified the manipulation was live, and sequence-alignment analysis found a small improvement on complete context-listed phrases, too small to materially change side-level WER, and for Gemini coexisting with worsened unlisted-token error. Evaluating context mechanisms therefore requires sequence-aligned term-level, insertion, and speaker-label measures alongside aggregate accuracy.
\end{abstract}

\section{Introduction}

A practitioner adapting automatic speech recognition (ASR) to a specialized domain now has an inexpensive option that did not exist a few years ago, which is to supply domain knowledge in the prompt. Modern LLM-based transcription APIs accept free-text conditioning. Intuition suggests that names, biographies, and curated vocabulary supplied at inference time should steer the model toward the right entities and conventions. Earlier results on prompt-conditioned Whisper support that intuition, with domain prompts roughly halving word error rate (WER) on air-traffic-control audio \citep{cochran24}. Production transcription tools ship the mechanism as a first-class feature.

This work asks whether the mechanism works in one deployment where it actually ships. The system under study is Dialog Scribe, a production oral-history transcription tool. Its single operator, coauthor Keith Nore, is an experienced volunteer transcriptionist, and he processed a private collection of 1970s--80s Skagway, Alaska oral-history cassettes through it. He attached rich per-project context to essentially every project: interviewer and interviewee names, a biography line, and a standard 34-entry vocabulary of local places, people, and railroad terms. His hand-corrected verbatim transcripts provide the references. The retained production audio and the application's own prompt-rendering code path make a controlled re-run possible, which is what allows the deployment itself to be studied rather than a reconstruction of it.

The design is a found-data, retrospective, paired within-item ablation. The same audio is reprocessed with the deployed transcription configuration held fixed within each contrast, only the prompt-level context varied, and each side-by-model cell scored against the same reference. The purpose of pairing within item was to eliminate between-arm differences in the composition of audio sides, and to control for each side's shared reference conventions and recording difficulty. Two properties make the result unusually defensible for a null. First, it is preregistered: hypotheses, metric definitions, exclusions, and the statistical plan were frozen before the complete outcome dataset existed in analyzed form, two disclosed gpt-4o pilot sides having been scored during scorer development, with the analysis code committed by SHA-256 hash (OSF 10.17605/OSF.IO/NS49B). The purpose of freezing the code, and not the plan alone, was to remove any post-hoc latitude in how the registered metrics were computed. Second, it is audited. The null was unexpected, so the manipulation was traced end to end after scoring. Prompt construction was byte-identical to the production code path, delivery was verified in both providers' request paths, and model consumption was evidenced by context-only tokens demonstrably present in the full-arm outputs and absent otherwise. The first objection to any null, that the treatment never reached the subject, is substantially ruled out by direct evidence.

The confirmatory answer is that prompt-level context produced no detectable change in aggregate accuracy on this corpus. None of the four registered hypotheses (WER, entity recall, full-vs-structure, over-biasing) was supported for either model. For \texttt{gpt-4o-transcribe} the interval is narrow for the observed single-pass outputs, in that side-level resampling of them concentrates the median paired difference within approximately one WER point of zero. That is not a repeated-call equivalence bound, as the post-hoc rerun demonstrates (\S\ref{sec:rerun}). It is also post-hoc rather than a registered equivalence test, and the sides cluster within eight interviewees (\S\ref{sec:threats}, T7 and T8). The exploratory answer is that the practitioner's real lever is elsewhere, since which deployed configuration is run moves WER by about 18 points on this audio. In the no-prompt single pass, the weaker configuration additionally produced looping or otherwise high-error output exceeding 100\% WER on 5 of the 19 degraded-cassette sides, against 1 of 19 for the stronger one (\S\ref{sec:exploratory}).

Contributions:

\begin{enumerate}
\item A preregistered, implementation-audited null: prompt-level context conditioning produced no detectable change in aggregate transcription accuracy on a production-derived corpus, for two deployed commercial transcription configurations. The gpt-4o bootstrap interval, about $\pm 1$ WER point under side-level resampling of the observed single-pass outputs, was part of the registered plan. Its interpretation as a practical sensitivity bound is post hoc, since no equivalence margin was preregistered.
\item An exploratory comparison of the two deployed provider configurations, with a reliability dimension: a median 17.8-point WER gap on the same audio, robust to reference-anchoring stratification, together with a 5-of-19 pathological-output rate for the weaker configuration in the no-prompt single pass, against 1 of 19 for the stronger one. The pipelines differ in more than model identity (\S\ref{sec:exploratory}). For this corpus, configuration selection and output screening appeared to offer substantially more operational leverage than prompt engineering.
\item A construct-validity account of why the null coexists with real context effects. The context demonstrably increases production of context-listed strings and elicits named speaker labels, with lower sequence-aligned error concentrated on complete listed phrases: for \texttt{gpt-4o-transcribe} under every construct variant tested, appearing across a majority of sides, remaining negative under every leave-one-interviewee-out deletion, and not driven by the three most frequent vocabulary items (\S\ref{sec:posthoc}); for Gemini only under the phrase-span construct, alongside worsened unlisted-token error. Those effects are arithmetically too small, or too entangled with output instability, to register in side-level WER, and they are diluted in generic entity-recall metrics. Evaluations of context mechanisms need sequence-aligned token-level constructs, such as listed-term error scored against the context's own term list and attribution accuracy, as first-class outcomes.
\item A documented workflow for evaluating null results in ASR: preregistration with frozen code, a post-hoc manipulation audit, and a threats-to-validity analysis organized around the burden inversion a null demands.
\end{enumerate}

The rest of this paper is organized as follows. Section~\ref{sec:related} reviews prompt conditioning, contextual biasing, and beyond-WER evaluation. Section~\ref{sec:system} describes the system under study, and Section~\ref{sec:method} the method, including the registered hypotheses and the frozen outcome measures. Section~\ref{sec:results} presents the confirmatory, sensitivity, exploratory, and labelled post-hoc results, and Section~\ref{sec:audit} the implementation audit. Section~\ref{sec:discussion} discusses where the operational leverage lies and what the null implies for evaluation design. Section~\ref{sec:threats} presents the threats-to-validity analysis, Section~\ref{sec:future} the limitations and future work, Section~\ref{sec:ethics} the ethics and data-governance record, and Section~\ref{sec:repro} the reproducibility and availability statement. Section~\ref{sec:conclusion} concludes.

\section{Background and Related Work}
\label{sec:related}

\subsection{Prompt conditioning for ASR}

Free-text conditioning of transcription is best described not as an established adaptation technique but as a widely productized mechanism with a recent, thin, and uneven evidence base. Its provenance is telling. Whisper's previous-text conditioning was introduced as a long-form decoding heuristic (the preceding window's transcript is fed to the decoder, and disabled when decoding is unreliable because it propagates errors), not as a domain-adaptation channel \citep{radford23}. OpenAI's hedges are scoped to that family: \texttt{whisper-1} prompting is ``more limited than our other language models'' \citep{openai26a}, and the Cookbook's caution that prompt techniques are ``not especially reliable'' concerns fictitious-spelling prompts for Whisper \citep{openai26b}. For the \texttt{gpt-4o-transcribe} family, by contrast, the same guide presents the \texttt{prompt} parameter as a working context channel, used ``similarly to how you would prompt other GPT-4o models'' \citep{openai26a}. The mechanism tested here is therefore vendor-endorsed for the exact model tested, which sharpens the question. Where the literature reports clear prompt-driven gains, the mechanism is usually something stronger than plain inference-time text: task-token engineering for unseen tasks \citep{peng23}, paired speech-text exemplars in the decoder context \citep{wang24}, dedicated prompt encoders cross-attending text into the speech encoder \citep{yang24b}, or fine-tuning the model to attend to prompts at all. Both \citet{liao23} and \citet{shamsian24} are premised on the observation that off-the-shelf Whisper does not reliably exploit textual prompts. Direct tests of plain-text prompting are mixed: \citet{gao25} report large overall gains on child read speech from prompted Whisper-plus-LLM configurations (9.4\% to 5.1\% WER for their best system), alongside a counterintuitive sensitivity in which prompts derived from the text being read worsened WER while irrelevant text improved it; \citet{yang24a} find no correlation between Whisper's prompt understanding and its accuracy; ProfASR-Bench reports ``little to no change'' in average WER even under oracle prompts, naming the phenomenon a \emph{context-utilization gap} \citep{piskala25}; ContextASR-Bench evaluates context exploitation at named-entity scale (300k+ entities, 10+ domains) and finds it highly model-dependent \citep{wang25}; and IndicContextEval, evaluating related commercial AudioLLM endpoints (\texttt{gpt-4o-transcribe} and Gemini 3 Flash), finds entity-list context moves \texttt{gpt-4o-transcribe} by only ${\sim}2.6$ WER points (with larger relative gains confined to entity error) and degrades one open model outright \citep{joshi26}. Against this, an earlier air-traffic-control result found domain prompts roughly halving Whisper Small and Medium WER in a vocabulary-saturated regime \citep{cochran24}. Section~\ref{sec:discussion} reconciles the two. What the present study adds is a preregistered, implementation-audited test of the exact shipped mechanism, on a production-derived corpus, through the production code path.

\subsection{Contextual biasing}

A distinct and far more mature family conditions the decoder rather than the instruction channel, with context supplied as a structured, bounded list. The line runs from on-the-fly WFST biasing in a production recognizer \citep{aleksic15} through subword shallow fusion for end-to-end models \citep{zhao19}, attention over an embedded phrase list (CLAS) \citep{pundak18}, trie-based deep biasing \citep{le21}, tree-constrained pointer generators \citep{sun21}, and contextual adapters over frozen models \citep{sathyendra22}; it carries over to Whisper-class models \citep{sun23} and scales via retrieval to 200k-entry inventories \citep{gong25}. Its known pathology, over-biasing (false insertion of list items absent from the audio), is documented from the beginning and actively engineered against \citep{alon19,zhao19}. It motivates the H4 arm. Several commercial services expose structured phrase, vocabulary, or keyterm interfaces distinct from unrestricted free-text domain descriptions, although their proprietary internal mechanisms are not always documented: Google's phrase sets with tunable boost, whose documentation states the false-positive tradeoff outright \citep{googlecloud26}, AWS custom vocabularies \citep{aws26}, AssemblyAI, which exposes both natural-language contextual prompting and an explicit keyterms list and whose current streaming documentation describes the two as complementary and usable together \citep{assemblyai26}, and Deepgram keyterm prompting \citep{deepgram26}. The null reported here concerns the prompt channel only and does not bear on decode-time biasing. In fact it motivates it (\S\ref{sec:discussion}).

\subsection{Beyond-WER evaluation}

That WER misaligns with downstream utility is a twenty-year-old observation \citep{wang03,favre13}, and benchmark WERs are known to flatter real-world performance \citep{szymanski20}. Proposed remedies make meaning and entities first-class: semantic-distance metrics that track human judgment better than WER \citep{kim21,kim22}, entity-aware measures from ATENE \citep{benjannet15} to entity-dense benchmarks \citep{delrio21}, the biasing literature's own B-WER/U-WER split, which scores list words separately from all others \citep{le21}, and, closest to the construct argument developed here, Contextual Earnings-22, which scores custom-vocabulary performance as its own outcome alongside aggregate accuracy for both keyword prompting and keyword boosting on realistic earnings-call audio \citep{durmus26}. Speaker attribution is likewise scored as its own construct in the meeting-transcription literature (cpWER) \citep{watanabe20}, and hallucinated content in foundation-model transcription is a documented failure mode in its own right \citep{koenecke24}, consistent with the pathological outputs observed here (\S\ref{sec:exploratory}). The results reported below give this literature a new exhibit: a context mechanism whose real effects are entirely in constructs that WER excludes or dilutes (\S\ref{sec:audit}, \S\ref{sec:threats} T11--T13).

\subsection{Preregistered evaluation and null results}

Preregistration separates confirmatory from exploratory claims and makes nulls publishable evidence rather than file-drawer residue \citep{nosek18}; the practice has been urged for NLP specifically \citep{vanmiltenburg21}, though its fit to exploratory research remains debated \citep{sogaard23}. The practice is followed strictly here (frozen hypotheses, frozen analysis code by hash, disclosed pilot exposure, a deviations policy), with one component added that computational studies particularly need: a post-hoc \emph{implementation audit} establishing that the manipulation physically occurred. The same methodological line appears in a preregistered comparison of retrieval architectures, which likewise found that instrument constructs, not headline metrics, carried the interesting variance \citep{cochran26}.

\section{System Under Study}
\label{sec:system}

\subsection{The tool}

Dialog Scribe is a production web application for interview and oral-history transcription. For each project the operator may attach \emph{context}: interviewer and interviewee names, a short biography, a curated custom vocabulary, and a domain template (here: \texttt{history}). At transcription time \texttt{DomainTemplateService.render\_prompt} composes these into a prompt. The OpenAI provider passes it as the API \texttt{prompt} parameter on every call (including every sub-chunk when large files are size-split), and the Gemini provider appends it as a \texttt{Context:} block inside its fixed instruction wrapper.

\subsection{The corpus}

The production database holds one operator's sustained real workload: 251 projects (an application ``project'' is one uploaded audio part, \S\ref{sec:sample}), ${\sim}40$ hours of audio, 331 transcriptions. The recordings are digitized 1970s--80s cassette interviews from a single private Skagway oral-history collection: one interviewer's voice throughout, eight interviewees, degraded consumer-era tape audio. Context attachment is near-universal (250 of 251 projects carry vocabulary, names, and bios), which is precisely why the context question cannot be answered observationally from this corpus. There is no context-off population, and the repeat transcriptions that exist vary model, not context. The study therefore re-runs the corpus rather than mining it.

\subsection{Reference transcripts (``gold'')}

The operator produced verbatim references at cassette-side granularity by exporting machine drafts and hand-correcting them externally (a provenance fact whose consequences are analyzed as threat T4). Corrections were extensive: reference-vs-machine-draft WER runs 15--50\%. ``Gold'' is used here as a term of art for the reference standard, and not as a claim of error-free ground truth (\S\ref{sec:threats}, T12).

\section{Method}
\label{sec:method}

\subsection{Design}

The design is a paired within-item ablation with two fully crossed factors, context arm (3 levels) and model (2 levels), applied to each of 19 cassette sides. The same audio and the same reference are used in every cell of a side. The inferential quantity is therefore the within-(side $\times$ model) paired difference between arms, which eliminates between-arm differences in side composition and controls for each side's shared reference conventions, recording difficulty, and provider wrappers (prompt effects may still interact with difficulty; the contrasts hold the items fixed rather than assuming additivity). This is a retrospective, found-data design and not a prospective randomized study. Much of what randomization would normally buy (exchangeability across arms on item difficulty) the pairing supplies by construction, while what it does not supply (arm execution order, provider-side drift across the batch window) is treated as a threat in \S\ref{sec:threats} (T5).

\subsection{Sample and pre-specified exclusions}
\label{sec:sample}

The confirmatory sample is a production-derived sample and not the full production database. At the freeze these 21 were every cassette side for which a complete side-level corrected transcript existed, and no study-specific sampling was applied on top of that. Membership therefore follows reference availability, which followed the operator's ordinary correction queue rather than any rule created for this study. The queue is not assumed random: it may itself have been shaped by interest, cassette order, audio quality, or the quality of the machine draft, none of which was recorded. Its formation is as follows:

\begin{center}
\begin{tabular}{lr}
\toprule
Stage & Count \\
\midrule
Part-projects (uploaded audio parts) in the production database & 251 (${\sim}40$ h audio, 331 transcriptions) \\
Cassette sides with operator-corrected verbatim references & 21 \\
Structurally excluded before scoring (see below) & 2 \\
\textbf{Confirmatory sample} & \textbf{19 sides $\cdot$ 70 audio parts $\cdot$ ${\sim}10.6$ h} \\
Part-level API records (70 parts $\times$ 2 models $\times$ 3 arms) & 420 \\
Physical provider calls behind those records (\S\ref{sec:conditions}) & 1{,}290 (OpenAI 1{,}080; Gemini 210) \\
\bottomrule
\end{tabular}
\end{center}

The unit hierarchy runs from physical medium to scored text as follows: collection $\rightarrow$ physical cassette $\rightarrow$ cassette side (the reference unit and the experimental unit) $\rightarrow$ uploaded audio part (${\sim}10$ minutes), which is the application's ``project'' (each upload creates its own project row carrying its own operator-entered context) $\rightarrow$ provider call (Gemini: one whole-file call per part; OpenAI: one call per size-split sub-chunk, 1--7 per part, \S\ref{sec:conditions}) $\rightarrow$ side-level hypothesis (the concatenation of the side's part transcripts; 19 sides $\times$ 2 models $\times$ 3 arms $=$ 114). The production database's 251 ``projects'' are therefore uploaded parts, not cassettes; ${\sim}40$ hours of audio corresponds to roughly 240 ten-minute parts. Because context attaches at the part-project level and the operator entered it per upload, a side's parts carried near-identical but not always byte-identical context. On 6 of the 19 sides, the parts' vocabulary lists or name and biography fields differ slightly (a term added or dropped, a second interviewee listed on some parts, wording variants; verified against the database snapshot). The full arm sent each part exactly its own production context, so these variations are production-faithful and constant across arms, and the 912--1245-character prompt range in \S\ref{sec:conditions} reflects them.

Interviewee composition is 5, 4, 2, 2, 2, 2, 1, 1 sides across eight interviewees. The two excluded sides were removed before scoring by structural inspection (one side's reference spans more audio than its labeled parts; another absorbs two mislabeled parts from a different side); both exclusions are registered, outcome-blind, and re-enterable only as a labelled secondary analysis after repair. The sample was fixed, with no optional stopping.

\subsection{Conditions}
\label{sec:conditions}

The model is held fixed within every contrast. The arms were generated against the application's own render path, and are called none, structure, and full throughout this paper (code identifiers in parentheses):

\begin{itemize}
\item \textbf{none} (\texttt{off\_generic}): no prompt at all (\texttt{prompt=None}).
\item \textbf{structure} (\texttt{off\_structure}): the general domain template plus generic speaker/format instructions (417 characters), with no names, biography, or vocabulary.
\item \textbf{full} (\texttt{on}): exactly what production sends, i.e.\ the project's \texttt{history} domain template, named-speaker instructions, biography, and the full curated vocabulary (912--1245 characters per project).
\end{itemize}

The two models are \texttt{gpt-4o-transcribe} (OpenAI; files $>$25 MB size-split into sub-chunks, prompt repeated per sub-chunk) and \texttt{gemini-2.5-flash} (Google; whole-file, no seams).

\subsubsection*{Call census and seam handling}

The 420 part-level records correspond to 1{,}290 physical provider calls: Gemini received one whole-file call per record (210 calls), while OpenAI received one call per sub-chunk (1{,}080 calls; 1--7 sub-chunks per part, six for 44 of the 70 parts). Sub-chunking is duration-proportional, targeting 20 MB per chunk (a buffer under the API's 25 MB limit). Each chunk after the first is given a 1-second audio lead-in so words are not lost at the boundary, and chunk transcripts are joined with a single space. Overlap duplicates are not removed, so up to about a second of speech can be transcribed twice at a seam, and no seam-specific normalization is applied before scoring. Chunk boundaries are identical across arms for every part (verified from the run log), so seams cannot confound the within-model arm contrasts; they do enter the exploratory cross-provider comparison (\S\ref{sec:exploratory}), where they work against the provider they burden. The same prompt is repeated on every sub-chunk. There is no per-chunk retry: a failed sub-chunk fails its whole part-record, and a rerun repeats the entire part with identical prompt and parameters. The retained run log shows a single uninterrupted pass (420 of 420 records, monotonic timestamps), so no cell was rerun.

\subsection{Outcome measures (frozen)}

All arms and gold were normalized identically (lowercase; straighten quotes; strip leading speaker labels; drop bracketed stage directions; strip punctuation except intra-word apostrophes; collapse whitespace). WER and CER were computed with \texttt{jiwer} \citep{vaessen25} on side-level text (the original run's environment file was not retained; rescoring under jiwer 4.0.0 reproduces the frozen per-cell output file byte-exactly, i.e.\ the serialized \texttt{score\_cells.csv} is byte-identical under file diff). Entity recall (alignment-based) is defined over gold proper-noun tokens (curated vocabulary plus mid-sentence-capitalized gold words, with capitalization identified in the original reference text before the case-folding normalization) as the fraction aligned as \texttt{equal}. Over-biasing (\texttt{ent\_over}) is the summed positive excess of each curated-vocabulary term's count in the hypothesis over its count in the gold. The metric code was frozen by SHA-256 before scoring.

\subsection{Hypotheses and statistical plan (as registered)}

\begin{itemize}
\item \textbf{H1 (primary):} full context reduces WER: paired $\Delta$ WER (full $-$ none) $< 0$, within each model.
\item \textbf{H2:} full context increases entity recall, within each model.
\item \textbf{H3 (full vs structure):} $\Delta$ WER (full $-$ structure) $< 0$. Registered as isolating the knowledge component (names, biography, vocabulary) from task structure; as \S\ref{sec:audit} details, the contrast additionally includes the history-vs-general template difference, so it compares the full production prompt against a generic structural prompt rather than surgically isolating knowledge.
\item \textbf{H4 (over-biasing):} context introduces excess injected-vocabulary tokens (\texttt{ent\_over} higher under full).
\end{itemize}

Hypotheses were registered with directional expectations; the registered tests are two-sided at $\alpha = 0.05$, with direction reported alongside. The registered test is a Wilcoxon signed-rank test \citep{wilcoxon45} on per-side paired differences, per model ($n = 19$), with zero differences dropped (the \texttt{wilcox} rule) and SciPy's automatic exact/normal selection. Per-contrast sign counts and Hodges--Lehmann estimates are reported in \S\ref{sec:posthoc}. Holm correction \citep{holm79} is applied across the two primary tests and within the H2--H4 family. Effects are reported as median paired $\Delta$ with 10{,}000-resample bootstrap percentile 95\% CIs, resampling sides as units. The test and the headline estimate are therefore not quite the same statistical object: the signed-rank test is naturally read as a test of a symmetric location shift, that is of the pseudomedian, whereas the bootstrap quantity is a separately estimated descriptive median. The Hodges--Lehmann estimates in \S\ref{sec:posthoc} are the estimate matched to the test, and they are reported alongside for that reason. The distinction matters most for gemini, whose difference distribution is heavy-tailed and asymmetric. A pre-specified sensitivity analysis excludes the two pilot sides scored during scorer development (disclosed in the registration; their Gemini arms were never inspected). Exploratory analyses (model interaction, interviewee stratification, chunk-length interaction, CER) were registered as hypothesis-generating only.

\subsection{Preregistration and timeline}
\label{sec:timeline}

The plan was committed to version control and registered (OSF 10.17605/OSF.IO/NS49B) before \texttt{score.py} ran on the full 2$\times$3 dataset; outcome data for the 17 non-pilot sides and the entire Gemini pass did not exist in analyzed form at freeze time. Analysis scripts are frozen by SHA-256 (hashes in the registration); any post-freeze change is a reportable deviation. Deviations: none. Analyses added in revision (\S\ref{sec:posthoc}) are labelled post-hoc additions computed outside the frozen scripts, not changes to them.

\subsubsection*{Execution record}

The full 2$\times$3 batch ran on 2026-07-11 (UTC): the OpenAI pass 00:12--01:50, the Gemini pass 01:50--03:34. Within each audio part the three arms ran adjacently in a fixed order (none $\rightarrow$ structure $\rightarrow$ full, typically within about 90 seconds), so provider-side drift would have to act within about a minute, per item, to confound arms. The order was fixed rather than counterbalanced (\S\ref{sec:threats}, T5). The runner logged prompt character counts rather than full prompt text, and it did not capture provider-reported version identifiers (the endpoints are aliases). Both are audit-logging lessons now specified for the production system.

\section{Results}
\label{sec:results}

\subsection{Confirmatory results}

\begin{center}
\begin{tabular}{@{}l p{2.5cm} p{4.3cm} p{4.3cm} l@{}}
\toprule
Hypothesis & Metric, contrast & \texttt{gpt-4o-transcribe} & \texttt{gemini-2.5-flash} & Verdict \\
\midrule
\textbf{H1} (primary) & WER, full $-$ none & \textbf{$+0.6$ pt} $[-1.1, +1.0]$ ($p = 0.57$; Holm $p = 1.0$) & \textbf{$-0.4$ pt} $[-2.5, +7.0]$ ($p = 1.0$; Holm $p = 1.0$) & not supported \\
\addlinespace
H2 & entity recall, full $-$ none & $-0.3$ pt $[-1.2, +1.6]$ ($p = 0.98$; Holm $p = 1.0$) & $-1.7$ pt $[-5.5, +1.2]$ ($p = 0.14$; Holm $p = 0.87$) & not supported \\
\addlinespace
H3 & WER, full $-$ structure & $+0.1$ pt $[-1.0, +0.9]$ ($p = 0.77$; Holm $p = 1.0$) & $+0.4$ pt $[-5.5, +10.1]$ ($p = 0.71$; Holm $p = 1.0$) & not supported \\
\addlinespace
H4 & over-biasing, full $-$ none & 0 tokens $[-1, +1]$ ($p = 0.79$; Holm $p = 1.0$) & $+1$ token $[-16, +7]$ ($p = 0.81$; Holm $p = 1.0$) & not supported \\
\bottomrule
\end{tabular}
\end{center}

\noindent\textbf{H1 with the two pilot sides excluded ($n = 17$): gpt-4o $+0.6$ pt ($p = 0.55$); gemini $-0.6$ pt ($p = 0.61$).} This registered sensitivity analysis is stated beside H1, and not only in \S\ref{sec:sensitivity}, because it is the check on the one part of the sample that was not outcome-blind at freeze. Two sides had been scored during scorer development; the other 17 sides' outcomes and the entire Gemini pass did not exist in analyzed form when the plan was registered (\S\ref{sec:timeline}).

\noindent\footnotesize\emph{Cells are median paired per-side $\Delta$ ($n = 19$ per model) with 10{,}000-resample bootstrap percentile 95\% CIs (fixed seed); WER and entity recall in percentage points, over-biasing in excess token counts. Holm correction within family (H1 $\times$ 2 models; H2--H4 $\times$ 2 models).}\normalsize

For gpt-4o the H1 interval is tight ($[-1.1, +1.0]$ WER points), so the result is informative at the side level: under the registered side-resampling analysis of the observed single-pass outputs, the 95\% interval did not extend beyond approximately one WER point in either direction. Two bounds attach to that statement. First, the bootstrap interval was part of the registered analysis plan, but its interpretation as a practical sensitivity bound is post hoc, because no equivalence margin or smallest effect size of interest was preregistered (\S\ref{sec:threats}, T7). Second, it resamples sides as independent units although the sides cluster within eight interviewees (\S\ref{sec:threats}, T8). The labelled post-hoc cluster-resampled sensitivity analysis (\S\ref{sec:posthoc}) produced a similarly small interval for gpt-4o on this observed set of clusters, and an essentially uninformative one for gemini. For gemini the interval is wide ($[-2.5, +7.0]$) even at the side level and its null is correspondingly weaker, so the headline claim rests on the gpt-4o bound (\S\ref{sec:threats}, T7). H4's null adds a bounded reassurance: no systematic increase was detected in the registered excess-count metric, though the gemini estimate is imprecise and the metric does not capture all forms of context-induced error (misplaced or substituted listed terms, damage to neighboring words).

\subsection{Pre-registered sensitivity}
\label{sec:sensitivity}

Excluding the two pilot sides did not materially change the H1 estimates: gpt-4o $+0.6$ pt ($p = 0.55$), gemini $-0.6$ pt ($p = 0.61$).

\subsection{Exploratory (hypothesis-generating; not multiple-comparison controlled)}
\label{sec:exploratory}

\begin{itemize}
\item The deployed configurations split widely. Paired per-side in the no-prompt arm, the \texttt{gpt-4o-transcribe} configuration beats the \texttt{gemini-2.5-flash} configuration by a median 17.8 WER points ($p < 0.0001$), and the gap is similar in the other arms (structure $+17.4$, full $+21.8$, both $p < 0.002$). This is a comparison of two deployed provider pipelines and not of models in isolation: the two paths differ in file segmentation (OpenAI size-splits; Gemini takes whole files), instruction wrapper, and endpoint type (a dedicated transcription endpoint versus a general multimodal model), so the gap should not be attributed to model identity alone. Because each reference was corrected from a machine draft anchored to one model (11 gpt-4o-anchored sides, 8 gemini-anchored), the comparison is stratified: the gap is a median 18.8 points on gpt-4o-anchored sides and 12.8 points on gemini-anchored sides, double digits even where the reference is anchored to the other model (\S\ref{sec:threats}, T4b).
\item A reliability gap compounds the accuracy gap. In the no-prompt arm, gpt-4o has a median WER of 35\% (IQR 20--45) with 1 of 19 sides pathological (WER $>$ 100\%), and gemini a median of 44\% (IQR 36--94) with 5 of 19 pathological. WER above 100\% establishes very high error-plus-insertion mass, and not by itself hallucination or looping. It is used here as an operational screen, and \S\ref{sec:posthoc} characterizes the flagged cells by hypothesis-to-reference length ratio (gemini 1.4--4.2; gpt-4o 1.1--1.2). The post-hoc inspection below characterizes every flagged cell. At these counts the contrast is descriptive, and no inferential claim is attached to it.
\item No chunk-length interaction was detected, over a narrow range. Spearman correlation between mean part length and per-side $\Delta$ WER is ${\approx}0$ for both models ($p > 0.96$), but nearly all parts are about 10 minutes, so the data cannot estimate an interaction beyond that range.
\item CER agrees with WER (H1 not supported on CER for either model), so the WER result is not a tokenization artifact.
\item Interviewee-level heterogeneity is unresolved: per-interviewee median $\Delta$ WER spans roughly $-10$ to $+11$ points at per-interviewee $n$ of 1--5 sides. This inspection is inconclusive rather than negative, and the spread is one reason cluster-aware uncertainty (\S\ref{sec:threats}, T8) matters.
\end{itemize}

\subsubsection*{Post-hoc inspection of high-WER cells (2026-07-19, after the confirmatory and rerun analyses)}

One author inspected all 12 flagged main-run cells with model and arm visible; no second adjudicator was used. Repetition statistics (most frequent exact contiguous 6- and 12-token sequences after the study normalization) were computed by script, and excerpts of each flagged output were read against the reference. Verbatim looping was found in 8 of 12 cells, all gemini: a single repeated passage recurring, per cell, between 46 and 3{,}924 times (these counts are observations from the inspected outputs, not a prespecified threshold). The four non-looping flagged cells (the three gpt-4o arms and gemini's full arm on one severely degraded side) contain locally fluent, thematically related text that nonetheless aligns to the reference almost nowhere (length ratios 1.0--1.5). Whether that text is supported by the audio was not adjudicated against the recordings, so it is not called hallucination here. The inspection script and its pseudonymized per-cell output are deposited with the release materials (\S\ref{sec:repro}).

\subsection{Labelled post-hoc additions (added during manuscript revision)}
\label{sec:posthoc}

\emph{Everything in this subsection was computed during manuscript revision (from 2026-07-15), outside the frozen pipeline (one labelled script, fixed seed 20260715, 10{,}000 resamples), on the frozen per-cell outputs and retained run texts. Nothing here alters a registered metric or test.}

\subsubsection*{Cluster-aware uncertainty (T8)}

An interviewee-cluster bootstrap (resampling the eight interviewee clusters with replacement) on the H1 paired differences gives gpt-4o a 95\% CI of $[-0.1, +1.0]$ WER points (identical under a seven-cluster family-clustering sensitivity variant that merges the two same-family interviewees; whether family membership is the right dependence structure for ASR errors is plausible but not established). For gemini it gives $[-25.5, +14.6]$ ($[-50.7, +15.3]$ under the family merge). This is a sensitivity analysis, not a population-level equivalence result: it produced a similarly small interval for this observed set of interviewee clusters, but the estimate is unstable with only eight unevenly represented clusters. The counterintuitive narrowing relative to the side-level interval ($[-1.1, +1.0]$) should not be read as stronger confirmation: the sides that improved under context concentrate in a few interviewees, and with eight clusters percentile-bootstrap bounds are highly discrete and sample-dependent. It should be noted that cluster resampling is still side-weighted (an interviewee with five sides contributes more whenever selected); it addresses within-interviewee dependence, not the interviewee-level estimand, which the next paragraph takes up. For gemini, whose pathological sides cluster within interviewees, the cluster-aware interval is essentially uninformative: its null should be read as ``no evidence,'' not ``evidence of no effect.'' H2--H4 cluster intervals all continue to span zero.

\subsubsection*{Interviewee-weighted effect (the estimand fix)}

Giving each interviewee equal weight (one median paired $\Delta$WER per interviewee, computed post hoc) yields eight summaries per model. For gpt-4o they have a median of $+0.6$ points (range $-10.3$ to $+11.0$), five positive and three negative. An exact sign test over the eight values gives $p = 0.727$. With eight interviewees the inference is necessarily coarse, but the interviewee-weighted summary is consistent with the side-level null. For gemini the eight summaries span $-233$ to $+50$ points, dominated by pathological sides concentrated in single-side interviewees; no useful interviewee-level inference is available.

\subsubsection*{Test detail}

H1 sign counts (positive $=$ context worse): gpt-4o 12/7/0 ($+$/$-$/0), Hodges--Lehmann $+0.4$ points; gemini 9/10/0, Hodges--Lehmann $+0.0$. Zero differences dropped per the \texttt{wilcox} rule (H2/H4 have 2--3 ties); SciPy auto-selects exact vs normal approximation.

\subsubsection*{Arm-level distributions}

\begin{figure}[htbp]
\centering
\includegraphics[width=\textwidth]{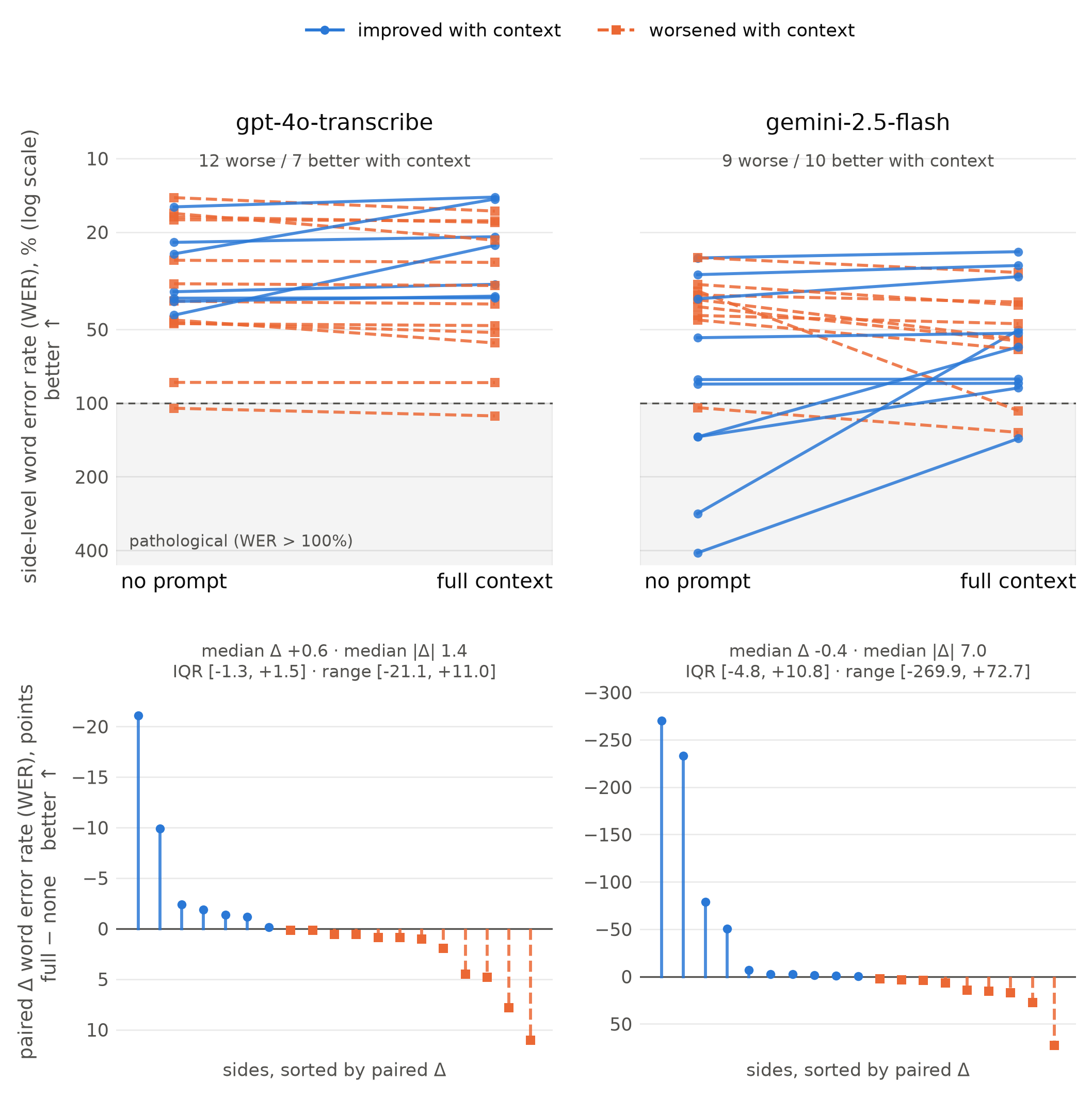}
\caption{Both vertical axes are inverted, so improvement points upward, and color, marker, and line style encode it as well. Top: all 19 sides' paired WER, no prompt $\rightarrow$ full context, per deployed configuration (log scale; shaded band $=$ pathological, WER $>$ 100\%). Bottom: the same paired differences (full $-$ none), sorted, on a linear point scale, which is the direct estimand. \texttt{gpt-4o-transcribe} changes were mixed and centered near zero (12 worse / 7 better; median $\Delta$ $+0.6$ points; median $|\Delta|$ 1.4; signed-$\Delta$ IQR $[-1.3, +1.5]$), although several sides showed larger individual changes (signed-$\Delta$ range $-21.1$ to $+11.0$). \texttt{gemini-2.5-flash} is dominated by arm-unstable high-WER outputs (median $|\Delta|$ 7.0, range $-269.9$ to $+72.7$): three of its five no-prompt-pathological sides recover under full context, one improves but remains pathological, one worsens within the band, and one previously healthy side crosses in (the arm-instability and corpus-weighted divergence reported above).}
\label{fig:paired}
\end{figure}

WER median [IQR] per arm: gpt-4o none 35.1 [19.9, 44.7], structure 34.6 [17.8, 46.9], full 32.7 [19.6, 43.8]; gemini none 43.9 [35.5, 94.0], structure 61.6 [37.7, 77.4], full 54.7 [39.2, 81.4]. Note the marginal medians and the registered paired median answer different questions: gpt-4o's full-arm marginal median is 2.4 points below its none-arm median while the median of its per-side paired differences is $+0.6$, which is what per-side heterogeneity looks like at $n = 19$. Figure~\ref{fig:paired} shows all 19 paired observations per configuration.

\subsubsection*{Side- vs corpus-weighted effects}

The registered estimand weights sides equally. Corpus-weighted (every reference word equal), $\Delta$ WER (full $-$ none) is $-0.4$ points for gpt-4o, against the registered side-weighted median of $+0.6$ points; the two are different estimands, and no interval from one bounds the other, but both are numerically close to zero. For gemini the corpus-weighted figure is $-9.5$ points, a divergence driven entirely by pathological no-prompt sides partially recovering under prompts (one side falls from 410\% to 140\% WER). That is a prompt-pathology interaction in an unstable configuration, not a lexical-accuracy gain, and it is invisible to the rank-based registered analysis by design (T9).

\subsubsection*{Listed-term coverage and alignment (the construct upgrade, T11b)}

The headline analysis is a B-WER/U-WER-style split \citep{le21}: over the \texttt{jiwer} word alignment of each side, every reference-token position is classified once as listed (a token of the side's curated vocabulary) or unlisted, and scored as correct iff its alignment operation is \texttt{equal}. The listed set covers 5{,}067 unique reference-token positions (4.6\% of ${\sim}110{,}400$); the unlisted complement covers 105{,}335.

\begin{center}
\begin{tabular}{@{}lcc@{}}
\toprule
& sequence-aligned listed-token & sequence-aligned unlisted-token \\
model & error, none $\rightarrow$ full & error, none $\rightarrow$ full \\
\midrule
\texttt{gpt-4o-transcribe} & 23.7\% $\rightarrow$ \textbf{22.0\%} & 34.9\% $\rightarrow$ 35.1\% \\
\texttt{gemini-2.5-flash} & 24.4\% $\rightarrow$ \textbf{25.8\%} & 36.3\% $\rightarrow$ 41.0\% \\
\bottomrule
\end{tabular}
\end{center}

For gpt-4o, full context produces a small sequence-aligned improvement concentrated exactly where the prompt aims (listed-token error falls 1.7 points while unlisted error is flat). For gemini, both rates worsen under full context, the unlisted rate sharply (its pathological, looping outputs concentrate there).

A complementary count-based census, bag-of-phrases occurrence coverage (per listed term: hypothesis occurrences capped at the gold count; a count metric, not a recall, since it is position-blind), of every curated-vocabulary term on every side (abbreviation-dot fragments excluded; 961 gold phrase occurrences corpus-wide):

\begin{center}
\begin{tabular}{@{}lcccc@{}}
\toprule
model & covered, none & covered, full & net $\Delta$ covered & excess listed-term occurrences none $\rightarrow$ full \\
\midrule
\texttt{gpt-4o-transcribe} & 690 (71.8\%) & 765 (79.6\%) & \textbf{$+75$ ($+7.8$ pp)} & 18 $\rightarrow$ 48 \\
\texttt{gemini-2.5-flash} & 647 (67.3\%) & 787 (81.9\%) & \textbf{$+140$ ($+14.6$ pp)} & 83 $\rightarrow$ 72 \\
\bottomrule
\end{tabular}
\end{center}

Corpus-wide top count-based movers: ``Whitehorse'' 25 $\rightarrow$ 98 of 121 gold occurrences (gemini); ``White Pass'' 173 $\rightarrow$ 193 of 211 and ``Skagway'' 177 $\rightarrow$ 194 of 227 (gpt-4o); ``Dyea'' 0 $\rightarrow$ 5 of 11 (gpt-4o). Harms exist: gemini's covered count for ``Bennett'' (Lake Bennett, a place on the railroad route, not a surname) falls 12 $\rightarrow$ 2 under full context. (The count-excess statistic is arm-comparative occurrence excess over the gold count, not positional insertion error; it cannot detect a listed string produced in the wrong place while another is missed.) Read jointly with the aligned split above and the sensitivity grid below, the two configurations' stories diverge. For gpt-4o, the count gains come with aligned improvement under every construct variant tested. For gemini, the larger count gain ($+14.6$ pp) coexists with lower sequence-aligned error on complete listed-phrase reference spans, alongside excess, unaligned production of listed strings (including into looping passages) and worsened error everywhere else. That is a prompt-sensitive lexical effect entangled with output instability rather than a clean accuracy improvement.

Metric mechanics, disclosed: the sequence-aligned metric's 5{,}067-token denominator consists of unique reference-token positions, each classified exactly once by token-set membership, so nested phrases cannot double-count it. The 961 phrase counts do count nested phrases per term (``White Pass'' inside the railroad's full name contributes to both), identically across arms. Matching is exact token-sequence after the study normalization (case and punctuation folded; no stemming, so plural or spelling variants do not match); each side's term list is its first part-project's vocabulary (near-identical across a side's parts, \S\ref{sec:sample}; the sensitivity grid below varies this). The aggregate arithmetic confirms the invisibility account with measured numbers, and legitimately so, because the sequence-aligned denominator is position-unique: gpt-4o's 1.7-point improvement across reference positions representing 4.6\% of the corpus has a corresponding reference-token component of ${\approx}0.08$ WER points (${\approx}0.07$ under the complete-phrase construct below), an order of magnitude below the typical scale of the observed per-side paired WER differences. Net WER additionally reflects insertions and alignment interactions on unlisted tokens, which the full WER computation captures but this component does not. The token-level response to context is now measured rather than merely illustrated: for gpt-4o it is a small sequence-aligned improvement on domain-salient terms, robust across constructs; for gemini it combines sequence-aligned gains confined to complete listed phrases with increased, often unaligned production of listed strings. Side-level WER does not isolate these prompt-specific lexical behaviors and, for the small gpt-4o gain, is too coarse to register their practical magnitude.

\subsubsection*{Construct sensitivity of the sequence-aligned metric (labelled post-hoc addition, 2026-07-19)}

Two construct choices above were flagged in review: the term list comes from each side's first part-project although the parsed term lists differ across parts on five of the 19 sides (\S\ref{sec:sample} reports six sides with any context variation, including name/biography fields), and ``listed'' membership is by bare token, so a component word of a multiword term counts as listed outside its phrase (``Pass'' without ``White Pass''). Because the references are side-level documents, a per-part recompute is not possible; instead a full sensitivity grid (labelled script \texttt{review3\_sensitivity.py}) recomputes the sequence-aligned split under three vocabulary definitions (first-part, union-of-parts, intersection-of-parts) crossed with two membership rules (token-set, as above; phrase-span, classifying only reference positions covered by a complete term occurrence, 1{,}271--1{,}323 positions or ${\sim}1.2\%$ of gold tokens versus 4{,}778--5{,}135 under token-set). The gpt-4o improvement survives every variant: $-1.5$ to $-2.4$ pp under token-set and larger, $-5.4$ to $-6.2$ pp (e.g.\ 29.0\% $\rightarrow$ 23.0\%, first-part definition), under phrase-span. For gemini the direction is construct-dependent, which sharpens its story: on complete listed phrases its sequence-aligned error also improves under full context ($-7.3$ to $-7.9$ pp), while its bare component tokens worsen slightly ($+0.2$ to $+1.4$ pp) and its unlisted error worsens sharply (above): lower sequence-aligned error on the exact phrase spans the prompt lists, coexisting with destabilization of the rest of the output. The arithmetic conclusion is unchanged under the stricter construct: a $-6$ pp change on ${\sim}1.2\%$ of gold tokens is ${\approx}0.07$ WER points. These are corpus-weighted descriptive rates; consistency across construct definitions does not by itself establish that the effect is distributed across sides, interviewees, or vocabulary items, which the distribution analysis below addresses. Two bounds on what this grid establishes: the three vocabulary definitions test robustness to plausible side-level constructions, but none exactly reconstructs part-specific prompt exposure, because the reference transcripts exist only at side level; and the analysis is sequence-aligned (a global textual edit alignment), not time-aligned, so under the complete-phrase construct the alignment assigns fewer substitution and deletion errors to listed phrase spans without proving that every credited occurrence was produced at the correct acoustic location, a caveat that matters most for gemini's looping outputs. A third bound is a dependency rather than a construct choice, and it is worth naming as reference--vocabulary coupling: the operator who curated the custom vocabulary is the operator who produced the reference transcripts, so the exact spelling of a rare term and the set of occurrences counted as correct both trace to the source that defined the prompt list. That does not change the fact that every arm is scored against the same reference, but the shared conventions can favor an output that reproduces the operator's own vocabulary and spellings, and the full-context arm is the arm prompted with them; the direction of that potential anchoring is the subject of T4. It also bounds the listed-phrase result, whose denominator and spellings are not independent of the manipulation.

\subsubsection*{Distribution of the complete-phrase result (labelled post-hoc addition, 2026-07-19)}

Because the grid's rates are corpus-weighted, they could in principle reflect a few long sides or high-frequency terms. A distribution check (labelled script \texttt{review5\_robustness.py}; phrase-span membership, first-part vocabulary; descriptive, no significance testing) finds the gpt-4o change broad rather than concentrated: per-side sequence-aligned listed-phrase error fell on 13 of 19 sides, rose on 4, and was unchanged on 2 (per-side median $-5.9$ pp); the corpus-weighted $-6.0$ pp changes only to between $-7.7$ and $-4.5$ pp under leave-one-interviewee-out deletion across all eight clusters; and excluding the three most frequent terms (``Skagway'' 227, ``White Pass'' 211, ``Whitehorse'' 121 gold occurrences) moves it to $-6.6$ pp, i.e.\ the result does not depend on the common local terms. Lower phrase-span error also occurred on 15 of 19 gemini sides (median $-9.3$ pp; leave-one-out range $[-9.7, -6.9]$; $-11.4$ pp excluding the top three terms), subject to the same sequence-alignment caveat and its instability elsewhere.

\subsubsection*{Pathology census}

Cells above 100\% WER: gemini 9 of 57 (5 sides in the none arm, 1 in structure, 3 in full; hypothesis/reference length ratios 1.4--4.2), gpt-4o 3 of 57 (one side, all three arms; ratios 1.1--1.2). Gemini's pathological set varies by arm, so the classification is arm-unstable; \S\ref{sec:rerun} measures its rep-to-rep stability directly.

\subsection{Labelled post-hoc addition: endpoint-stochasticity rerun (2026-07-18)}
\label{sec:rerun}

\emph{Run and analyzed during manuscript revision, outside the frozen pipeline (labelled scripts \texttt{rerun\_stochastic.py} and \texttt{score\_rerun.py}, run-level seed 20260718). Design fixed during manuscript revision, before the run: a subset of sides spanning the observed difficulty range, both deployed configurations, all three prompt arms, five repetitions per cell, with per-(part, model, repetition) counterbalanced arm order (seeded, logged per record). Nothing here alters a registered metric or test.}

Seven days after the main run (rerun 2026-07-18 UTC, 21:11--23:45; main batch 2026-07-11 UTC, \S\ref{sec:timeline}), four sides were reprocessed through the same production code path: one low-WER side, one moderate, one side pathological for gemini in the none arm only, and one pathological for both models (single-pass WERs 16\% to 410\%). The rendered prompts were length-identical to the main run in all 33 (part $\times$ arm) cells, so the code path was unchanged. The 330 part-level records completed without error, corresponding to 990 physical provider calls (OpenAI: 825 sub-chunk calls, with each part's chunk count, 3--6, constant across arms and repetitions; Gemini: 165 whole-file calls; census as in \S\ref{sec:conditions}). They yielded 120 scored (side $\times$ model $\times$ arm $\times$ repetition) cells under the frozen scoring functions (the rerun scorer reproduces the frozen per-cell WERs of the main run exactly, 24/24 overlapping cells).

Within-arm rerun variability is small where behavior is stable and heavy-tailed where it is not. For \texttt{gpt-4o-transcribe}, rep-to-rep WER standard deviations were 0.04--0.1 points on the moderate side and 0.5--5 points elsewhere, but individual cells produced occasional large draws (one structure-arm repetition at 77\% WER against a 42\% median; one none-arm repetition at 28\% against an 18\% median). For gemini, the stable sides behaved like gpt-4o (SD 0.3--0.6 points on the moderate side) while the two unstable sides were extreme: none-arm SDs of 167 and 80 points, with single draws from 39\% to 525\% WER on the same audio under the same prompt.

The observed gpt-4o prompt difference is smaller than run-to-run pipeline variability on the rerun subset. In the original run, the four selected sides' own full-minus-none differences were $+7.8$, $+1.0$, $+0.2$, and $-1.4$ points, median $+0.6$, coincidentally matching the $+0.6$ registered median of the complete 19-side corpus (the subset summary is post hoc; only the 19-side analysis is registered). Across the 20 rerun (side $\times$ repetition) pairs, the paired $\Delta$WER (full $-$ none) had median $+0.2$ points, IQR $[-1.9, +2.7]$, range $[-10.2, +25.0]$. Differences of the magnitude observed in both the subset and the complete corpus are therefore small relative to run-to-run variability on these selected sides: such effects cannot be distinguished reliably from pipeline stochasticity using one realized transcription per cell. This does not independently establish equivalence, and it does not strengthen the original side-resampled interval as a bound on repeated-run expected performance; it shows that the confirmatory result should be read as a finite-corpus, single-pass finding whose effect size lies below the run-to-run resolution measured on these sides, while the rerun does not estimate that resolution across the complete 19-side corpus.

Pathological output is a stochastic, arm-dependent rate, not a stable side property. The both-models-pathological side was pathological in 14 of 15 gpt-4o cells and in 5/5 gemini none-arm repetitions, but only 2/5 under structure and 2/5 under full context. The gemini-only side was pathological in 1/5 none-arm and 1/5 structure-arm repetitions and 0/5 under full context: the main run's 283\% WER draw there was an unlucky draw from a bimodal distribution whose rerun median is 41\%. Single-pass pathology counts (the ``5 of 19 sides'' census) should therefore be read as draw-dependent; per-arm pathological-draw counts provide a more informative description than a binary classification of each side.

On the two unstable sides, full context was associated with lower observed gemini dispersion and fewer pathological draws. The complete per-cell dispersion statistics (five repetitions each; robust measures reported because $n = 5$ and the tails are heavy):

\begin{center}
\scriptsize
\setlength{\tabcolsep}{3pt}
\begin{tabular}{@{}ll rrc rrc@{}}
\toprule
& & \multicolumn{3}{c}{\texttt{gpt-4o-transcribe}} & \multicolumn{3}{c}{\texttt{gemini-2.5-flash}} \\
\cmidrule(lr){3-5}\cmidrule(lr){6-8}
Side & Arm & median [IQR] (MAD) & min--max & path/5 & median [IQR] (MAD) & min--max & path/5 \\
\midrule
S03 & none & 107.1 [105.0, 108.8] (2.1) & 105.0--109.3 & 5/5 & 285.0 [241.9, 503.3] (140.8) & 144.2--524.6 & 5/5 \\
S03 & structure & 104.2 [103.6, 106.6] (2.4) & 100.3--109.6 & 5/5 & 94.5 [93.7, 570.6] (1.1) & 93.4--576.0 & 2/5 \\
S03 & full & 103.9 [100.1, 105.2] (3.7) & 99.2--108.1 & 4/5 & 99.9 [99.9, 133.6] (3.0) & 96.9--139.8 & 2/5 \\
\addlinespace
S06 & none & 37.8 [37.4, 38.6] (0.8) & 36.7--40.9 & 0/5 & 41.3 [40.8, 53.0] (2.2) & 39.1--222.9 & 1/5 \\
S06 & structure & 41.6 [39.8, 48.7] (5.0) & 36.6--77.2 & 0/5 & 43.4 [40.1, 63.0] (5.8) & 37.6--342.3 & 1/5 \\
S06 & full & 41.3 [39.7, 50.4] (2.1) & 39.2--65.9 & 0/5 & 55.1 [48.9, 65.9] (9.5) & 45.6--71.6 & 0/5 \\
\addlinespace
S11 & none & 82.4 [82.4, 82.5] (0.1) & 82.3--82.6 & 0/5 & 83.0 [83.0, 83.1] (0.0) & 82.8--84.1 & 0/5 \\
S11 & structure & 82.4 [82.4, 82.5] (0.1) & 82.4--82.6 & 0/5 & 83.2 [83.1, 83.8] (0.1) & 83.1--84.6 & 0/5 \\
S11 & full & 82.5 [82.5, 82.5] (0.0) & 82.5--82.6 & 0/5 & 82.8 [82.6, 82.9] (0.2) & 82.5--83.3 & 0/5 \\
\addlinespace
S18 & none & 18.1 [16.2, 21.1] (2.4) & 15.7--28.0 & 0/5 & 29.3 [28.2, 29.9] (1.1) & 26.8--36.0 & 0/5 \\
S18 & structure & 16.1 [15.8, 16.5] (0.3) & 15.7--16.8 & 0/5 & 29.2 [28.6, 30.0] (0.8) & 28.4--71.5 & 0/5 \\
S18 & full & 16.5 [16.2, 16.5] (0.3) & 14.1--17.8 & 0/5 & 30.1 [29.0, 35.3] (2.6) & 27.5--49.7 & 0/5 \\
\bottomrule
\end{tabular}
\normalsize
\end{center}

\noindent\footnotesize\emph{Rerun WER dispersion per (side $\times$ arm $\times$ configuration): median [IQR] (median absolute deviation), min--max, and pathological draws (WER $>$ 100\%) out of five, all in WER points; side IDs are the release pseudonyms.}\normalsize

On S03 (pathological for both configurations) the gemini none arm's dispersion collapses under full context (MAD 140.8 $\rightarrow$ 3.0, range 144--525 $\rightarrow$ 97--140, pathological draws 5/5 $\rightarrow$ 2/5); on S06 the extreme tail disappears (max 222.9 $\rightarrow$ 71.6, 1/5 $\rightarrow$ 0/5 pathological) while median WER rises (41.3 $\rightarrow$ 55.1). The structure arm produced the largest single blowups for both configurations (576\% and 342\% for gemini, 77\% for gpt-4o). This suggests a possible stabilization effect: full context suppressed gemini's extreme draws on the two deliberately selected unstable sides, at the cost of a higher median on one of them. Resting as it does on two unstable sides, five repetitions per cell, post-hoc side selection, and unversioned model aliases, it requires confirmation on a larger, prospectively selected sample. It does establish that rerun noise is not symmetric across arms; for gpt-4o no consistent arm-variance ordering emerged.

The sequence-aligned listed-term measure was stable on the rerun subset; the subset carries little of the corpus-level effect. On these four sides the main run showed a full-vs-none sequence-aligned listed-token error difference of $-0.3$ pp for gpt-4o (18.6\% $\rightarrow$ 18.3\%; the corpus-level $-1.7$ pp of \S\ref{sec:posthoc} is concentrated in other sides). The rerun reproduced this subset value in every repetition (per-repetition paired differences $-0.3$ to $+0.2$ pp), so the measure itself is stable under endpoint stochasticity; whether the corpus-level $-1.7$ pp would reproduce is not answered by this subset. Gemini's listed-term error varied widely across repetitions (15.0\% to 30.3\% in the full arm), consistent with its instability elsewhere.

\subsubsection*{Containment check}

19 of the 24 main-run cells fall within their five-repetition rerun ranges; the five exceptions are two trivial exceedances ($\le 0.2$ points) and three draws in the heavy-tailed cells (by 5--60 points), where five repetitions evidently do not bracket the tail. Median rerun WERs match the main-run values closely in all stable cells.

Caveats: the subset is four sides chosen post hoc on observed WER to span difficulty; seven days elapsed between runs and provider-side model aliases are unversioned, so main-run-vs-rerun differences confound time with draw; and all analyses here are descriptive, with no multiplicity control.

\section{The Implementation Audit}
\label{sec:audit}

A null invites one objection before all others: the context never reached the model. Because the result was unexpected, the pipeline was audited end to end after scoring, covering backend prompt construction, both providers' delivery paths, the database snapshot, the runner, and the run's logged output. The chain of evidence follows.

\begin{enumerate}
\item Arm construction is byte-identical to production. The runner's arms map onto the application's own code path: \texttt{off\_generic} $=$ \texttt{prompt=None}; \texttt{off\_structure} $=$ the general template with generic speaker instructions (417 chars); \texttt{on} $=$ exactly the production render for the project (all 70 parts used the \texttt{history} template; 912--1245 chars).
\item The context is real and substantial. A representative rendered prompt contains both participants' names, a biography line, and the operator's standard 34-entry vocabulary (local place names, historical figures, railroad terms) appended under \texttt{Important terms:}.
\item Delivery is verified in code. For OpenAI, \texttt{params["prompt"]} is set on every call, with the same prompt repeated on every size-split sub-chunk (no dilution across seams). For Gemini, context is appended inside the provider's fixed instruction wrapper, which is constant across arms and cancels in the within-model pairing.
\item The run's logged prompts are length-consistent with expected renders. The expected prompt for each of the 420 records was re-rendered from its part-project row (\S\ref{sec:sample}; importing the application's own template module against the database snapshot) and compared with the prompt character counts the runner logged, with zero mismatches. No full-arm prompt was missing its vocabulary. The logged artifact is the character count, so this step verifies length consistency, not byte equality of the transmitted strings; byte-identity holds for the construction path (item 1). The runner did not retain full prompt text, so this length-level check is final; hash-level prompt logging is the corresponding production requirement.
\item The models demonstrably received and reacted to the context. On one interview's sides, tested at both ends of the prompt, speaker-name labels appear only in the full arm (gpt-4o 17/11 occurrences on the two sides; gemini 594/348), and rare listed terms appear only in the full arm (``hostler'': 0 $\rightarrow$ 1--2 occurrences in both models; gemini's recognition of ``Whitehorse'' doubles, 16 $\rightarrow$ 32). The label counts establish prompt-sensitive output change, not correct attribution: the labels were not scored against annotated speaker turns, and gemini's counts far exceed plausible turn counts for a two-voice interview, consistent with over-segmentation or repetition. Gemini's extreme counts are read here as prompt reactivity, potentially pathological, rather than as a produced benefit. Delivery ``verified in code'' (item 3) establishes what the request contained, not what the proprietary endpoint did internally; the output changes here are the evidence of consumption.
\end{enumerate}

\subsection{Mechanistic account of the null}

The context has two real effects. The first is speaker-label generation, which the normalization deliberately excludes from WER (it strips speaker labels from gold and hypothesis alike, because attribution is a usability construct rather than lexical accuracy; label correctness is unscored, \S\ref{sec:threats} T13). The second is listed-term production, which the post-hoc census (\S\ref{sec:posthoc}) measures at $+7.8$ to $+14.6$ percentage points of bag-of-phrases occurrence coverage, and of which the sequence-aligned share concentrates on complete listed phrases (robust across construct variants for gpt-4o; under the phrase-span construct only for gemini, \S\ref{sec:posthoc}). That share is arithmetically worth under 0.1 WER points, an order of magnitude below the typical scale of the observed per-side paired WER differences. A listed term can only help where it is both spoken and otherwise misheard, and on this audio that intersection covers a few hundred of about 110{,}000 gold tokens. The manipulation worked. Its most clearly observed lexical effects were too sparse to materially change side-level WER on this corpus. The audit's one-interview examples above generalize corpus-wide in \S\ref{sec:posthoc}'s systematic table.

The audit surfaced two as-registered design nuances, stated here for precision. H3's contrast is between a generic frame and everything the application sends, so it includes the history-vs-general template difference rather than a surgical isolation of the context sections (a factorial ablation separating template, names, biography, and vocabulary is future work). Gemini's none arm retains the provider's fixed wrapper (constant across arms). Neither changes the interpretation of the registered within-configuration contrasts, and none of the registered alternatives was supported.

\section{Discussion}
\label{sec:discussion}

For a practitioner transcribing archival oral history with current deployed transcription configurations, the two configurations differed much more in observed WER than the prompt arms did. The gap was a median of about 18 points between the two provider pipelines, and in the no-prompt single pass the weaker pipeline produced WER above 100\% on 5 of 19 sides against 1 of 19 for the other configuration, although the configuration comparison was exploratory and the pipelines were not matched (\S\ref{sec:exploratory}). The decision the tooling makes most visible, context prompting, produced no detectable aggregate effect. Investment in configuration bake-offs on representative audio, and in automatic pathological-output screening such as looping and repetition detectors and output-length bounds, is therefore the foremost operational recommendation here, ahead of prompt engineering. The rerun (\S\ref{sec:rerun}) sharpens that recommendation, since pathological output arrived stochastically per repetition rather than deterministically per input, so screening must run on every output. One further operational observation from \S\ref{sec:rerun} is deliberately narrow. On the two unstable sides included in the post-hoc rerun, full context was associated with lower observed gemini dispersion and fewer pathological draws. That is a possible stabilization effect, and it requires confirmation on a larger, prospectively selected sample before it can inform practice.

The manipulation tested is the prompt channel, which is the shipped mechanism. Decode-time contextual biasing (shallow fusion, CLAS-style architectures, retrieval-conditioned decoding) operates on the decoder's hypothesis space directly, and it is the natural next candidate for the domain-salient term gains this domain needs. This study detected no aggregate-level improvement from the prompt channel on this corpus, and thereby sharpens the case for a stronger or more direct contextual-biasing mechanism. It does not condemn context per se (\S\ref{sec:threats}, T16).

The audit and the post-hoc census show context producing measurable, prompt-specific output changes. Context elicited named speaker labels and increased bag-of-phrases listed-term coverage by 8 to 15 percentage points, with lower sequence-aligned error concentrated on complete listed phrases (\S\ref{sec:posthoc}). WER treats those two changes differently, and the difference is worth keeping separate: the speaker labels are excluded outright, because the study normalization strips them from reference and hypothesis alike, while the listed-term gains are scored but diluted, since the same analysis prices the sequence-aligned gain at under 0.1 WER points against a corpus of about 110{,}000 gold tokens. An evaluation that is intended to see what context does must therefore make sequence-aligned token-level constructs first-class: listed-term error scored against the context's own term list, and attribution accuracy as its own outcome. This is the study's most general contribution, in that two constructs can both be called ``accuracy'', be measured on the same output, and move independently. Aggregate accuracy metrics alone are insufficient for evaluating contextual mechanisms whose intended effects are concentrated in rare terms, entities, or speaker labeling.

The earlier ATC result \citep{cochran24} found prompts roughly halving WER on Whisper Small and Medium. Error sources were not formally categorized in either study, so the reconciliation offered here is a hypothesis rather than a demonstrated mechanism. A plausible explanation is that the ATC corpus placed a large fraction of its error mass on prompt-addressable vocabulary, such as callsigns and waypoints, with high prompt-reference overlap and with overlap correlating with gain, and that it used far smaller models. Errors in the present corpus, by contrast, appear to include substantial acoustic-degradation and long-form stability components that no listed term can address, and frontier-scale models already command the general vocabulary. The broader proposition, that prompt benefit scales with the prompt-addressable share of the error mass, is a useful hypothesis for future work rather than an established generalization. Recent purpose-built benchmarks land on the same side of the reconciliation. Even oracle prompts produce little aggregate-WER movement \citep{piskala25}, and entity-list gains on related commercial endpoints are small, concentrated in entity-level metrics, and model-dependent \citep{joshi26}.

\section{Threats to Validity}
\label{sec:threats}

\emph{Organized by the Shadish--Cook--Campbell taxonomy \citep{shadish02} rather than a flat limitations list, because a null result inverts the burden of proof: for a positive result, threats ask whether the effect is real; for a null, they ask whether a real effect could have been missed, shifting the weight onto statistical-conclusion and construct validity.}

\subsection{Internal validity: substantially mitigated by design}

\textbf{T1. Manipulation failure} (``the context never reached the model''). Substantially ruled out by the audit (\S\ref{sec:audit}): byte-identical prompt construction, 420/420 logged prompt lengths matching expected renders, and context-only tokens present in full-arm outputs at both ends of the prompt. Length matching cannot exclude every conceivable prompt corruption, but wholesale manipulation failure is strongly disfavored.

\textbf{T2. Selection bias} (operators reach for context on harder audio). Removed by design: every side contributes all three arms, so arm composition is identical and each side's difficulty is controlled by the pairing (though prompt effects may still interact with difficulty).

\textbf{T3. Model confound.} Removed by design for the confirmatory contrasts: model is held fixed within every contrast and fully crossed; both models ran in a single timestamped batch.

\textbf{T4. Reference anchoring} (confirmed, and conservative for the headline). The operator produced the gold by correcting exported machine drafts generated in production with context on, so the reference is anchored toward the full arm's output style. Two mitigations. Direction: because the references were corrected from full-context drafts, residual anchoring would be expected to favor the full arm's wording and spelling conventions, making a context benefit somewhat easier to observe; the direction is plausible rather than formally guaranteed, however, because the magnitude and structure of residual reference error (inherited spellings, omission patterns, segmentation conventions) are unknown. Magnitude: gold-vs-draft WER runs 15--50\%, so corrections were extensive and the reference diverges substantially from any draft.

\textbf{T4b. Model anchoring of the reference} (consequence for the \emph{exploratory} model finding). Each gold is anchored to whichever model produced the corrected draft (11 gpt-4o, 8 gemini). Checked by stratification: the paired model gap is a median 18.8 points on gpt-4o-anchored sides and 12.8 points on gemini-anchored sides. The smaller gap on gemini-anchored sides is consistent with anchoring contributing to the gap's size, but gpt-4o retains a double-digit advantage on the sides whose references were anchored to gemini drafts, so the configuration-gap finding survives stratification. Residual confounding by interviewee and audio quality across the two strata cannot be excluded at this $n$; the estimates are descriptive.

\textbf{T5. Order, carryover, and provider drift.} Carryover is excluded (independent, stateless API calls; fixed reference). Two residual temporal threats are not: arm execution order was fixed rather than randomized or counterbalanced, and a hosted endpoint can change provider-side without notice. The run log bounds both tightly (\S\ref{sec:timeline}): the whole batch ran inside 3.5 hours on 2026-07-11, and within each audio part the three arms executed adjacently (fixed order none $\rightarrow$ structure $\rightarrow$ full, typically within ${\sim}90$ seconds), so drift would have to act within about a minute, per item, and be arm-correlated, to bias the paired contrasts. The fixed order was not specified in the registration; it is disclosed here.

\textbf{T6. Attrition/exclusion bias.} The two excluded sides were removed by structural inspection before scoring, preregistered, outcome-blind by construction.

\subsection{Statistical-conclusion validity: the crux for a null}

\textbf{T7. Sensitivity (the null's central obligation).} A null must state what effect sizes the data speak against. For \texttt{gpt-4o-transcribe} the 95\% bootstrap CI on the median paired $\Delta$ WER is $[-1.1, +1.0]$ points: the registered side-resampling analysis of the observed single-pass outputs provides little support for a median side-level effect larger than about one point in either direction, under resampling that treats sides as independent units (see T8). For \texttt{gemini-2.5-flash} the CI is wide ($[-2.5, +7.0]$), so its negative result is materially weaker; the headline rests on the gpt-4o interval. The bootstrap interval was part of the registered analysis plan; its interpretation as a practical sensitivity bound is post hoc, because no equivalence margin or smallest effect size of interest was preregistered, and it targets the sample's side-level location statistic, not automatically a population-average effect. Because each cell was transcribed once, it describes the observed single-pass outputs, not the pipelines' repeated-run expectations (T8b); a TOST-style equivalence design \citep{lakens17} is the right upgrade for any follow-up.

\textbf{T8. Effective sample size, clustering, and estimand.} Nominally $n = 19$ per model, but the sides cluster: eight interviewees, one family contributing 9 of 19 sides, and one voice, the interviewer's, present on every recording. Effective $n$ is below 19, and any interviewer-specific interaction with context is unobservable. Two estimands must be distinguished: the finite-corpus question (what happened across these 19 sides), which the registered analysis answers, and the collection- or speaker-level question (what to expect on other sides or speakers from this domain), for which side-level resampling understates uncertainty. The labelled post-hoc cluster-resampled sensitivity analysis (\S\ref{sec:posthoc}) probes the dependence part: it produced a similarly small interval for gpt-4o ($[-0.1, +1.0]$) on this observed set of clusters and an essentially uninformative one for gemini ($[-25.5, +14.6]$); the interviewee-weighted summaries with an exact sign test (\S\ref{sec:posthoc}) address the estimand part and are likewise consistent with the gpt-4o null. With only eight clusters both are crude, and the safest reading of the null remains about the observed corpus, not an underlying speaker population.

\textbf{T8b. Endpoint stochasticity (one run per cell in the main study).} Each (side $\times$ model $\times$ arm) cell of the confirmatory analysis was transcribed once, and hosted generative endpoints are stochastic. The stratified rerun subset (\S\ref{sec:rerun}: four sides, both configurations, all arms, five repetitions, counterbalanced arm order, seven days later) bounds a source of measurement variability that the confirmatory analysis did not model: on those four sides, run-to-run pipeline variability exceeds the observed gpt-4o arm difference, so an effect of that size cannot be resolved from one realized transcription per cell. The subset's own single-pass median difference was $+0.6$ points, coincidentally equal to the median across the complete corpus (\S\ref{sec:rerun}); the variability itself was measured only on the four sides. Variability of this magnitude could equally have masked, exaggerated, or produced the small observed difference, and the gemini reruns show it is not symmetric across arms. The rerun therefore supports treating the confirmatory result as a finite-corpus, single-pass finding, and it weakens any reading of the side-resampled confidence interval (T7) as a bound on the pipelines' expected behavior over repeated runs: T7's interval describes the observed single-pass outputs. Further limits: the rerun covers four of 19 sides, its five repetitions do not bracket the tails of the unstable cells, and the seven-day gap means time and draw are confounded in main-run-vs-rerun comparisons. Separately, it establishes that gemini's pathological outputs recur as stochastic, arm-dependent rates rather than fixed side properties.

\textbf{T9. Outlier-driven noise} (gemini's five pathological sides). Handled by the registered analysis: rank-based Wilcoxon and median bootstraps are resistant to them, and the pathologies are reported as a reliability finding rather than silently absorbed.

\textbf{T10. Multiplicity and robustness.} Holm correction within registered families; none of the registered alternatives was supported. The pilot-excluded and CER analyses did not reverse the gpt-4o H1 result, while the gemini estimates remained too unstable for a strong negative inference. No single metric, test, or subset carries the conclusion.

\subsection{Construct validity: the deepest issue}

\textbf{T11. Aggregate metrics cannot see token-level effects (the central caveat).} The audit found context genuinely increasing production of listed rare terms and doubling recognition of a key place name: prompt-sensitive, domain-salient lexical changes amounting to a handful of tokens per 5--8k-word side (${\approx}0.02$--0.05 WER points on that per-side arithmetic, distinct from the corpus-level sequence-aligned component of \S\ref{sec:posthoc}), arithmetically invisible to aggregate WER and diluted even in our entity-recall metric, whose token set spans hundreds of proper nouns rather than the ${\sim}30$ listed terms. Whether these changes altered historical meaning or user utility was not measured; the terms are domain-salient, and we say no more than that. The claim must be bounded precisely: \textbf{prompt-level context did not move aggregate accuracy constructs here; it does not follow that context does nothing of domain value.} The same methodological moral appears in a preregistered comparison of retrieval architectures, where instrument constructs rather than headline metrics carried the interesting variance \citep{cochran26}: related constructs measured on the same output can move independently.

\textbf{T11b. The entity-recall construct is an approximation.} The registered entity metric's token set (curated vocabulary plus mid-sentence-capitalized gold words) is an ad hoc operationalization of ``entities'': a token-level recall over hundreds of proper-noun tokens, which dilutes any effect on the ${\sim}34$ prompted terms. The directly relevant metric, aligned listed-term error scored against the context's own term list, was not registered; \S\ref{sec:posthoc} computes it post hoc (labelled as such), and the follow-up design registers it as first-class (\S\ref{sec:future}). The \S\ref{sec:posthoc} count-vs-alignment divergence for gemini also shows directly that count-based term metrics can overstate contextual utility for unstable models; aligned scoring (the B-WER/U-WER-style split \S\ref{sec:posthoc} now reports) is the safer construct. Even the aligned construct is membership-rule-dependent: \S\ref{sec:posthoc}'s sensitivity grid shows gemini's aligned listed error \emph{worsening} under bare token-set membership and \emph{improving} under phrase-span membership, so evaluations should state and, where feasible, vary the membership rule.

\textbf{T12. What the reference measures.} The gold encodes one experienced transcriptionist's conventions (readability choices, rendering of dialect and false starts). Absolute WER against it conflates ASR error with those conventions, which is why the inferential quantity is the within-item paired $\Delta$, cancelling the shared offset. Absolute WER levels (15--50\%) should not be read as pure model error. No second transcriptionist independently validated the reference transcripts or adjudicated the post-hoc listed-term disagreements; consequently, rare-term spellings and the classification of contested occurrences depend on the same transcriptionist who created the references and curated the vocabulary. This does not invalidate the paired analysis, but it bounds the post-hoc rare-term results, which are the paper's most granular findings.

\textbf{T13. Deliberate construct exclusions.} The normalization strips speaker labels from gold and hypothesis alike, so the full arm's most visible output change, named speaker labels, is excluded from WER \emph{by design}. Attribution is a usability construct, not lexical accuracy; we report label generation descriptively (labels appear only under full context) and keep it out of the accuracy claim. Label \emph{correctness} was not scored: without turn-annotated references we cannot distinguish correct attribution from over-segmentation, repetition, or mislabeling, so the observation is that the prompt elicits labels, not that it attributes speech correctly.

\textbf{T14. Arm-composition nuances.} As registered and as production behaves: the H3 contrast includes the template difference (generic vs history frame), and gemini's none arm retains the provider's constant wrapper, so ``no prompt'' is not identical across providers. Both are stated so the construct each contrast operationalizes is precise; neither affects the within-model conclusions.

\subsection{External validity: bounded and pre-stated}
\label{sec:extvalidity}

\textbf{T15. Generalization bounds.} Single operator; single private collection; one interviewer's voice throughout; degraded 1970s--80s cassette audio; English; a ${\sim}10$-minute segmentation regime (so segment-length interactions were tested over a narrow range); two commercial models at a timestamped version, both subject to unannounced provider-side change. Reference availability was not randomly sampled either: the sides carrying a corrected transcript are those the operator had reached in an ordinary correction queue (\S\ref{sec:sample}), whose ordering was not recorded. None of these were sampled, so none support generalization; the registration stated this bound before the data existed.

\textbf{T16. Mechanism bound (the most important one).} The manipulation is \emph{prompt-level} context, the mechanism the deployed product ships. The result says nothing about decode-time biasing or retrieval-conditioned decoding, which operate on the decoder rather than the instruction channel. The null motivates exactly that stronger, more direct mechanism; it does not condemn context per se.

\textbf{Closing the loop.} The pattern across these threats is deliberate: the within-item design and the implementation audit address the internal-validity threats a null uniquely invites; the sensitivity reading converts ``not significant'' into ``the observed side-resampling analysis provides little support for a median side-level effect beyond about $\pm 1$ WER point'' for one model, with cluster-resampled and interviewee-weighted sensitivity analyses consistent with it; and the construct analysis bounds the claim to aggregate metrics while documenting the token-level effects those metrics cannot see. What survives is a precise negative: prompt-level context produced no detectable change in the registered side-level WER estimand in the observed single-pass corpus.

\section{Limitations and Future Work}
\label{sec:future}

Beyond the bounds in \S\ref{sec:extvalidity}, the two structurally excluded sides await repair and would enter only as a labelled secondary analysis. A follow-up designed as an equivalence test (TOST) with a pre-specified smallest effect size of interest would convert the sensitivity bound into a registered claim. Two next studies are natural. The first is a model bake-off on a second, governed archival collection with independently produced references (removing the reference-anchoring threat entirely), with sequence-aligned token-level constructs (listed-term error scored against the context's own term list; speaker-attribution accuracy) registered as first-class outcomes. The second is a decode-time biasing implementation evaluated on the same substrate, testing whether the mechanism the null motivates delivers the domain-salient term gains the prompt channel did not.

\section{Ethics and Data Governance}
\label{sec:ethics}

\subsection{Collection, ownership, and authorization}

The recordings are digitized 1970s--80s interviews from a single private oral-history collection, held and controlled by its original interviewer, who represented that he owns the collection and who is the one living voice on the recordings. The owner gave verbal permission for the collection's transcription and for this research use to coauthor Nore, the collection's sole operator in the system under study, who produced the production workload, its curated per-project context, and the verbatim reference transcripts. That permission covered processing through commercial transcription APIs under terms that exclude the submitted content from model training, which is the configuration described below. The owner re-confirmed it directly to the first author on 2026-08-28, before this manuscript was posted, covering the transcription, the research use, the commercial-API processing, and publication. The authorization is verbal throughout; no written instrument was executed, and the date recorded here is the authors' record of the conversation rather than a document signed by the owner. No consent or deposit agreements were made at recording time (common for private collections of this era), so no recorded term of collection speaks to computational processing or third-party services, and the authorization relied on is the owner's own, given with the processing described. Per the owner and operator, nothing in how the interviews were obtained conflicts with the processing described here. All interviewees are believed deceased. The owner reviewed the manuscript before posting and is not named in it, at the authors' choice and with his agreement.

\subsection{Authors' ethics and human-subjects assessment}

No IRB review or independent human-subjects determination was obtained. The first author's institution is an independent research nonprofit without an IRB. The authors considered the study to be secondary analysis of a private historical collection, under the U.S. Common Rule's definitions \citep{hhs18}, with OHRP's guidance on coded private information and biospecimens \citep{ohrp18} consulted for its treatment of secondary research, with no intervention or interaction with any individual. That assessment was not independently adjudicated. All recorded interviewees are believed deceased, which is a good-faith belief rather than a verified vital-records determination. The recordings nevertheless carry the identifiable voice of the collection's living owner and may contain information about living third parties, so the deceased-interviewee fact does not by itself settle the question. The owner authorized the transcription, the research use, and the commercial-API processing. This section therefore reports the relevant facts and safeguards, which are redaction of participant names and release of derived aggregates only, without asserting an independent regulatory determination.

\subsection{Provider data handling}

Audio was processed through two commercial transcription APIs, in the original production workload and again in this study's re-runs, under an organizational account held by the first author's institution in both cases. For OpenAI, audio was submitted inline to the \texttt{/v1/audio/transcriptions} endpoint under a project-scoped organizational key. Per OpenAI's endpoint-specific data-controls table \citep{openai26c} (archived captures of 2026-07-10 and 2026-07-21 bracket both runs, the intervening captures of 2026-07-11, 07-13 and 07-16 carry the same table, and the live page still did on 2026-08-28), API content on this endpoint is not used for model training absent an explicit opt-in (the organization's sharing opt-in is disabled), and the endpoint's listed default abuse-monitoring and application-state retention is ``None'' (the same page's general prose describes up-to-30-day abuse-monitoring logs for API usage generally; the endpoint table lists none for this endpoint). No separate zero-data-retention agreement was in place. For Google, processing used the Gemini Developer API (not Vertex AI) under a Cloud project with active billing, that is, Google's paid tier. Those terms \citep{google26a} (archived captures of 2026-07-10 and 2026-07-21 bracket both runs, the intervening captures of 2026-07-12 and 07-16 are identical, and so was the live page on 2026-08-28) state that prompts and responses are not used to improve Google products and are logged ``for a limited period of time'' solely for abuse monitoring. The terms do not specify the duration for ordinary generate-content calls (their explicit 30-day periods apply only to grounding features not used here). Audio was uploaded through the provider's Files API and deleted by the application immediately upon successful transcription, with the provider's 48-hour automatic file expiry \citep{google26b} bounding any failure path. Participant names, a biography line, and the curated vocabulary were transmitted to both providers by design in the full-context arm (\S\ref{sec:audit} demonstrates receipt). The unpaid tier of the Gemini Developer API carries materially different terms (content use for product improvement and human review). It did not apply here.

\subsection{Disclosure}

Participant names are redacted throughout this paper, matching the disclosure standard of the public preregistration; the collection's owner is likewise unnamed. The quoted curated-vocabulary examples and place names carry residual reidentification risk in combination (region, era, occupations, family structure). The first author reviewed and approved their publication under the preregistration's disclosure standard (2026-07-18), and the owner has since read the manuscript in which they appear (2026-08-28). All reported quantities are derived, non-identifying aggregates; the audio, database snapshot, and reference transcripts are not released (\S\ref{sec:repro}).

\section{Reproducibility and Data Availability}
\label{sec:repro}

The preregistration, including frozen hypotheses, metric definitions, statistical plan, and SHA-256 hashes of the four analysis scripts, is public (OSF 10.17605/OSF.IO/NS49B, registered from the study's OSF component, \url{https://osf.io/9nsvr}, which holds the release materials). Three of the four analysis scripts are attached to the registration and mirrored in the component's file storage with the public preregistration PDF. A release archive in the component's file storage (\texttt{context-effect-osf-release-2026-07-18.zip}; SHA-256 \texttt{366d7acfff7edeeefdd9a7f7991a376af911eb5723003cc7188693e50c9b6821}; 17 files) accompanies the registration, containing: the pseudonymized per-(side, model, arm) outcome file for the main study (114 rows) and the per-(side, model, arm, repetition) outcome file for the stochasticity rerun (120 rows), under one consistent pseudonym scheme (sides S01--S19, interviewee clusters C1--C8); a data dictionary; the labelled post-hoc analysis and figure scripts (seed 20260715) and the rerun runner and scorer (seed 20260718); complete console output of the confirmatory, post-hoc, and rerun analyses (side names pseudonymized, no other edits); software versions and a locked dependency file; and SHA-256 hashes of the retained raw artifacts. A second archive in the same component (\texttt{context-effect-provider-docs-2026-08-28.zip}; SHA-256 \texttt{05d58cbb5e6f7cdc474b0e86a0fcad9f5d3d1e78567407d111a516e0061ed95a}) holds dated snapshots of the commercial and agency documentation cited here, separating Internet Archive captures at or near the access dates from a same-day re-fetch used only as a drift check, with the SHA-256 of every file. It records that three of the four vendor documentation URLs cited in \S\ref{sec:related} had already moved by 2026-08-28, each still resolving by redirect. The fourth frozen script (\texttt{build\_join.py}) embeds participant surnames in its name-normalization tables; its full logic is released as \texttt{build\_join\_release.py} with the private name data externalized and a synthetic fixture (invented names) that exercises every join mechanism, and the original's freeze remains documented by hash. One release redaction is marked in-line in the post-hoc script (a family surname externalized to an environment variable). It is otherwise identical to the script that produced the reported numbers. The round-3 construct-sensitivity script (\texttt{review3\_sensitivity.py}, deterministic, no seed), the round-5 distribution-robustness and high-WER inspection scripts (\texttt{review5\_robustness.py}, \texttt{inspect\_flagged.py}), and the pseudonymized console outputs of the sensitivity grid, the rerun dispersion table, the distribution analysis, and the per-cell inspection record (\S\ref{sec:exploratory}--\S\ref{sec:rerun}) are deposited alongside the archive in the same component. The underlying audio, database snapshot, reference transcripts, raw model outputs, and pseudonym mapping contain personal information from a private collection and are not released (\S\ref{sec:ethics}). Their content is committed by hash in the archive, and all released quantities are derived, non-identifying aggregates. Provider version identifiers were not logged at run time (\S\ref{sec:timeline}) and cannot be added retroactively.

\section{Conclusion}
\label{sec:conclusion}

In a preregistered paired ablation on 19 cassette sides from one production-derived oral-history collection, full prompt-level context produced no detectable improvement in side-level WER. A post-hoc implementation audit verified at the prompt-construction and output-token level that the mechanism was live. The \texttt{gpt-4o-transcribe} differences were concentrated near zero in the observed sample, although the study was not designed as an equivalence test. In a post-hoc rerun of four deliberately selected sides, run-to-run pipeline variability exceeded the gpt-4o arm difference observed on those four sides, whose own single-pass median of $+0.6$ points coincidentally equalled the median across the complete corpus. On those selected sides, effects of that magnitude could not be resolved reliably from one realized transcription per cell. It should be noted that the rerun does not establish equivalence and does not provide a repeated-run population bound. The Gemini results were substantially less stable and do not support a comparable negative inference. The rerun showed the Gemini pathological behavior to be stochastic and arm-dependent, and on the two unstable sides full context was associated with fewer pathological draws and lower dispersion, a potential stabilization effect requiring prospective replication. Token-level analysis showed that the prompts did affect the outputs. Sequence-aligned error on complete listed phrases decreased for both configurations (for gpt-4o robustly across construct variants, on a majority of sides, under leave-one-interviewee-out deletion, and not driven by the most frequent terms). Gemini additionally produced more listed strings outside complete phrases, and with worsened unlisted-token error. Supplementing WER with sequence-aligned term-level, insertion, and speaker-label measures is therefore the foremost recommendation for evaluating contextual transcription mechanisms. For practitioners, the exploratory gap between the two deployed configurations, 17.8 median WER points and, in the no-prompt single pass, pathological output on 5 of 19 sides against 1 of 19, argues for representative pipeline evaluation and pathological-output screening ahead of prompt engineering. That gap should not be attributed to model identity alone. Because the rerun shows pathology striking stochastically per run, screening must apply to every output rather than to a blacklist of known-bad inputs. For the field, inexpensive preregistered nulls on shipped mechanisms remain worth running, because these mechanisms are widely deployed and their value remains sparsely measured in production-derived settings.

\section*{Author Contributions}

Cochran: conception, system, study design, preregistration, execution, analysis, audit, drafting (\S1--7, \S9--12). Dodson: threats-to-validity analysis and external-validity framing (\S8), methodological review. Nore: corpus creation (the production transcription workload and its curated context) and the verbatim reference transcripts.

\section*{Competing Interests}

Cochran developed Dialog Scribe, the production system whose shipped prompt-conditioning mechanism is the object of this study, and works with AI for Altruism, the nonprofit that operates it. He received no compensation for this study, which was carried out pro bono. Nore is the system's operator and produced both the corpus and the reference transcripts. Provider access was purchased at list price under an institutional account; neither OpenAI nor Google funded, reviewed, or otherwise supported this work, and no author holds a financial interest in either provider. Dodson declares no competing interests.

\bibliography{refs}

\end{document}